\documentclass{article}
\usepackage[preprint]{spconf}
\usepackage{amsmath,graphicx}
\usepackage[ruled,vlined]{algorithm2e}
\usepackage{lineno}
\usepackage{amssymb}
\usepackage{soul,color}
\usepackage{fancyhdr}

\newcommand\blfootnote[1]{%
  \begingroup
  \renewcommand\thefootnote{}\footnote{#1}%
  \addtocounter{footnote}{-1}%
  \endgroup
}

\fancypagestyle{firstpage}{%
  \lfoot{\fontsize{6}{6} \selectfont Copyright 2018 IEEE. Published in the IEEE 2018 International Conference on Image Processing (ICIP 2018), scheduled for 7-10 October 2018 in Athens, Greece. Personal use of this material is permitted. However, permission to reprint/republish this material for advertising or promotional purposes or for creating new collective works for resale or redistribution to servers or lists, or to reuse any copyrighted component of this work in other works, must be obtained from the IEEE. Contact: Manager, Copyrights and Permissions / IEEE Service Center / 445 Hoes Lane / P.O. Box 1331 / Piscataway, NJ 08855-1331, USA. Telephone: + Intl. 908-562-3966.}
 
}

\newcommand{\EQ}{\begin{equation}}
\newcommand{\NQ}{\end{equation}}
\newcommand{\ER}{\begin{eqnarray}}
\newcommand{\NR}{\end{eqnarray}}
\newcommand{\ERS}{\begin{eqnarray*}}
\newcommand{\NRS}{\end{eqnarray*}}
\newcommand{\bit}{\begin{itemize}}
\newcommand{\ben}{\begin{enumerate}}
\newcommand{\eben}{\end{enumerate}}
\newcommand{\ebit}{\end{itemize}}
\newcommand{\bzero}{{\bf 0}}
\newcommand{\ba}{{\bf a}}
\newcommand{\bb}{{\bf b}}
\newcommand{\bc}{{\bf c}}

\newcommand{\bx}{{\bf x}}
\newcommand{\by}{{\bf y}}

\newcommand{\bC}{{\bf C}}
\newcommand{\bD}{{\bf D}}

\newcommand{\bG}{{\bf G}}
\newcommand{\bH}{{\bf H}}

\newcommand{\bM}{{\bf M}}

\newcommand{\bS}{{\bf S}}

\newcommand{\bX}{{\bf X}}
\newcommand{\bY}{{\bf Y}}
\newcommand{\bZ}{{\bf Z}}

\title{UNSUPERVISED DOMAIN ADAPTATION USING REGULARIZED HYPER-GRAPH MATCHING}
\name{Debasmit Das \quad C.S. George Lee}
\address{ School of Electrical and Computer Engineering, Purdue University}
\begin{document}
\ninept
\thispagestyle{firstpage}
\maketitle
\begin{abstract}
Domain adaptation (DA) addresses the real-world image classification problem 
of discrepancy between training (source) and testing (target) data distributions.
We propose an unsupervised DA method
that considers the presence of only unlabelled data in the target domain. 
Our approach centers on finding matches between samples of the source and target domains. 
The matches are obtained by treating the source and target domains
as hyper-graphs and carrying out a class-regularized hyper-graph matching 
using first-, second- and third-order similarities between the graphs. 
We have also developed a computationally efficient algorithm by initially 
selecting a subset of the samples to construct a graph and 
then developing a customized optimization routine for graph-matching based on Conditional Gradient and Alternating Direction Multiplier Method. 
This allows the proposed method to be used widely. 
We also performed a set of experiments on
standard object recognition datasets to validate 
the effectiveness of our framework over previous approaches.
\end{abstract}
\blfootnote{This work was supported in part by the National Science Foundation under Grant IIS-1813935.
Any opinion, findings, and conclusions or recommendations expressed in this material are
those of the authors and do not necessarily reflect the views of the National Science Foundation.}
\section{Introduction}
\label{sec:intro}
\begin{figure}[h]
\centering
\includegraphics[width=8cm]{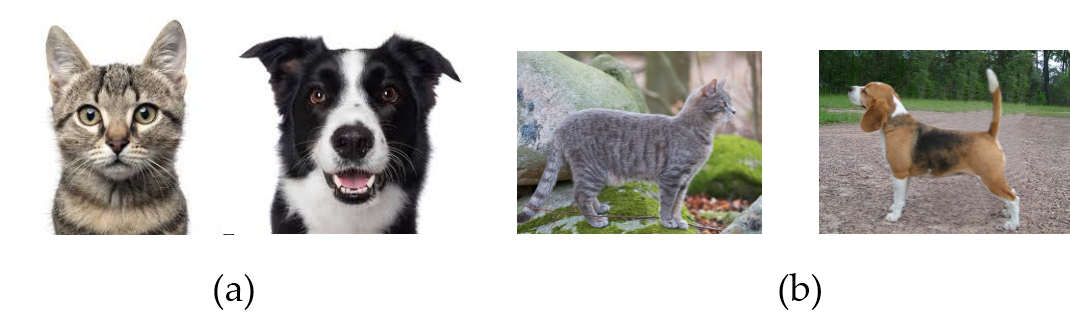}
\vspace*{-0.0in}
\caption{Source (a) and Target (b) domain has front and side views respectively.}
\label{cat-dog}
\vspace*{-0.2in}
\end{figure}

The assumption that test data is drawn from the same distribution 
as the training data is rarely encountered in real-world, machine learning problems. 
For example, consider a recognition system that distinguishes 
between a cat and a dog, and has been trained using labelled samples of the type 
shown in Fig.~\ref{cat-dog}(a). 
The same system when used to test in a different domain 
such as on the side images of cats and dogs (see Fig.~\ref{cat-dog}(b)) 
would fail miserably. 
This is because the recognition system has developed
a bias in only distinguishing between the face of a dog and a cat and not their side images. 
Domain adaptation (DA) aims to mitigate this dataset bias~\cite{torralba2011unbiased}, 
where different datasets have their own unique properties.
In this work, we consider \emph{unsupervised domain adaptation} (UDA) where we have labelled source domain data and only unlabelled target domain data.

Most previous UDA methods can be broadly classified into two categories. 
\emph{Instance Re-weighting} was one of the early methods, 
where it was assumed that conditional distributions 
were shared between the two domains. 
The instance re-weighting involved estimating 
the ratio between the likelihoods of being a source example or a target example 
to compute the weight of an instance. 
This was done by estimating the likelihoods independently~\cite{zadrozny2004learning} 
or by approximating the ratio between the densities~\cite{kanamori2009efficient,sugiyama2008direct}. 
One of the most popular measures used to weigh data instances 
was the Maximum Mean Discrepancy (MMD)~\cite{huang2007correcting}, 
which  computed the divergence between the data distributions in the two domains. 

The \emph{Feature Transfer} category consists of the bulk of the UDA methods. 
Well known feature transfer methods include  
the Geodesic Flow Sampling (GFS)~\cite{gopalan2014unsupervised,gopalan2011domain} 
and the Geodesic Flow Kernel (GFK)~\cite{gong2012geodesic,gong2013connecting}, 
where the domains are embedded in $d$-dimensional linear subspaces that can be seen as points on the Grassman manifold. 
The Subspace Alignment  (SA)~\cite{fernando2013unsupervised} 
learns an alignment between the source subspace 
obtained by Principal Component Analysis (PCA) and the target PCA subspace, 
where the PCA dimensions are selected by 
minimizing the Bregman divergence between the subspaces. 
Similarly, the linear Correlation Alignment (CORAL)~\cite{sun2016return} 
algorithm minimizes the domain shift using the covariance 
of the source and target distributions. 
Transfer Component Analysis (TCA)~\cite{pan2011domain} 
discovers common latent features having the same marginal distribution 
across the source and target domains. 
The Optimal Transport for Domain Adaptation ~\cite{courty2016optimal}, 
considers a local transportation plan for each source example to be mapped close to the target samples. 
Our approach is similar to~\cite{courty2016optimal} 
in the sense that it considers local sample-to-sample matching 
and this generally results in better performance than global methods 
because it considers the effect of each and every sample in the dataset explicitly. 
In~\cite{courty2016optimal}, 
they considered a first-order, point-wise unary cost between each source and target sample. 
Our approach develops a framework that exploit higher-order relations along with these unary relations. 
Such higher order relations provide additional geometric and structural information 
about the data beyond the unary point-wise relations. 
Hence, we expect higher order matching between source and target domains 
to yield better domain adaptation. 
In fact, a higher-order graph matching problem has been previously 
used for finding feature correspondences in images~\cite{duchenne2011tensor,le2017alternating} 
through a tensor-based
formulation but has not been applied to domain adaptation. 

\begin{figure}[]
\centering
\includegraphics[width=8cm]{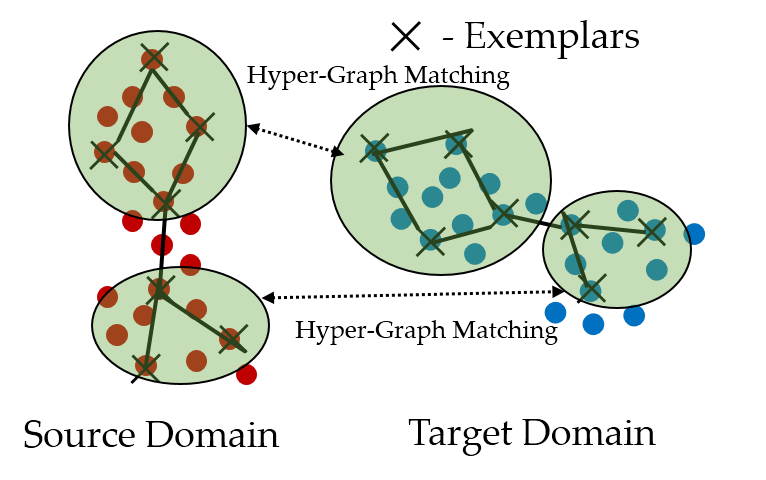}
\caption{Our approach involves extracting exemplars from both source and target domain. Hyper-graphs are constructed from those exemplars followed by matching.}
\label{graph}
\end{figure}

In that sense, our contributions are in the following ways:
(1) A mathematical framework 
using all the first-, second- and third-order relations 
to match the source- and target-domain samples 
along with a regularization using labels of the source-domain data. 
(2) Computationally efficient method of obtaining the solution of the optimization problem 
by solving a series of sub-problems using Alternating Direction Multiplier Method (ADMM). 
Moreover, we perform an initial clustering to select the most relevant instances and thus reduce 
the number of data points to be used in the optimization approach.
(3) Experimental evaluation on an object recognition dataset with analysis of the effect of each cost term. 
Fig.~\ref{graph} shows the intuition of our method in a two-dimensional setting. 
\section{PROPOSED DOMAIN-ADAPTATION METHOD}
\label{sec:pagestyle}
\subsection{Problem formulation}
For unsupervised domain adaptation (UDA), 
we have the source domain data matrix $\bX^s \in \mathbb{R}^{n_s \times d}$, 
vector of source domain data labels $\by^s \in \mathbb{R}^{n_s} $ 
and the target domain data matrix $\bX^t \in \mathbb{R}^{n_t \times d}$. 
Here, $n_s$ and $n_t$ are the number of source and target samples respectively, 
and $d$ is the dimension of the feature space. 
The labels in $\by^s$ range in $\{1,2, \cdots, C\}$. 
Both the domains share the same $C$ number of classes. For the purpose of UDA, we transform the source domain data close to the target domain data such that a classifier trained on the transformed source domain predicts well on the target instances.

\subsubsection{Finding Exemplars}
In our proposed method, 
we initially perform clustering to extract the set of exemplars, 
which is a representative subset of the original source and target domain dataset. 
This clustering is required to increase the computational efficiency 
of our hyper-graph matching method. 
We use Affinity Propagation (AP)~\cite{dueck2007non} to extract the exemplars. 
AP is an efficient clustering algorithm that uses \emph{message passing} between data-points. 
The algorithm requires the similarity matrix $\bS$ of a dataset as the input. 
Here, $[\bS]_{ij}=-||\bx_i-\bx_j||^2, i\neq j$ and $[\bS]_{ii}$ 
is the preference of sample $\bx_i$ to be an exemplar 
and is set to the same value of $p$ for all instances. 
$p$ controls the number of exemplars obtained. 
For low values of $p$ we obtain less exemplars 
while for large values of $p$ we obtain more exemplars. 
To obtain a desired number of exemplars, 
a bisection method that adjusts $p$ iteratively is used. 
For our purpose, 
we set $\eta$ to be the desired fraction of exemplars for 
both the source and the target domain dataset. 
Thus, as an output of the AP algorithm, 
we would obtain the exemplar source and 
target domain matrix $\bX'^{s} \in \mathbb{R}^{n'_s \times d}$ 
and $\bX'^{t} \in \mathbb{R}^{n'_t \times d}$, respectively, 
where $n'_s \approx \eta n_s$, $n'_t \approx \eta n_t$. 
However, to keep up with the notation, 
from now onwards in the paper we would denote $\bX^{s}$, $\bX^{t}$ 
as the exemplar source and target datasets and $n_s$ and $n_t$ 
as the number of source and target exemplars.

\subsubsection{Hyper-Graph Matching}
To carry out hyper-graph matching between the source and target exemplars, 
we consider all first-, second- and third-order matching between the source and target exemplars. 
We do not use just the higher order matching because the source 
and target hyper-graphs will be far from isomorphic 
and using only the structural information will produce misleading results. 
On the contrary for registration between similarly shaped objects~\cite{duchenne2011tensor}, using only
higher order matching produces excellent results.

We seek to find a matching matrix $\bC \in \mathbb{R}^{n_s \times n_t}$, 
where $[\bC]_{ij}$ is the measure of correspondence between source exemplar $i$ and target exemplar $j$.
For the first-order matching, 
we would like exemplars in the source domain 
to be close to similar exemplars in the target domain. 
Since the operation $\bC\bX^{t}$ rearranges the target exemplars, 
we would like the re-arranged target exemplars $\bC\bX^{t}$ 
to be close to $\bX^{s}$. 
Thus, to enable first-order matching, 
we would like to minimize the normalized term $f_1(\bC)=||\bC\bX^t-\bX^s||^2_\mathcal{F}/(n_sd)$, where $||\cdot||_\mathcal{F}$ is the Frobenius norm.

For the second-order matching, 
we would like source exemplar pairs to match with similar target exemplar pairs. 
This is carried out by initially constructing a source 
and a target adjacency matrix with source and target exemplars as graph nodes. 
If $\bD^s$ and $\bD^t$ are source and target adjacency matrices, 
then $[\bD^s]_{ij} = \text{exp}({-\frac{||\bx^s_i-\bx^s_j||^2_2}{{\sigma}_s^2}})$, $[\bD^t]_{ij} 
= \text{exp}({-\frac{||\bx^t_i-\bx^t_j||^2_2}{{\sigma}_t^2}})$ 
for $i \neq j$ and $[\bD^s]_{ii}=[\bD^t]_{ii}=0$. 
${\sigma}_s$ and ${\sigma}_t$ can be found heuristically 
as the mean sample-to-sample pairwise distance 
in the source and target domains, respectively.
For the second-order similarity, 
we want the row re-arranged target domain adjacency matrix $\bC\bD^t$ 
to be close to the column-rearranged source domain adjacency matrix $\bD^s\bC$. 
Taking into consideration the difference in the numbers of source 
and target exemplars $n_s$ and $n_t$, 
second-order graph matching implies minimizing 
the term $f_2(\bC)=||\bC\bD^t-r\bD^s\bC||^2_\mathcal{F}$, where $r=\frac{n_t}{n_s}$ is a correction factor. 

For the third-order matching problem, 
we use the tensor based cost term~\cite{duchenne2011tensor}. 
In that paper, they try to maximize the cost term 
$f_3(\bC)=\bH \otimes_1 \bc \otimes_2 \bc \otimes_3 \bc$, 
where $\bH \in \mathbb{R}^{{n_sn_t} \times {n_sn_t} \times {n_sn_t}}$ 
is a third-order tensor and the index $k$ in $\otimes_k$ indicates tensor multiplication 
on the $k^{th}$ dimension 
and $\bc \in \mathbb{R}^{n_sn_t}$ is the vectorized matching matrix $\bC$. 
Here, $[\bH]_{ijk}=\text{exp}(-\gamma||f_{i_s,j_s,k_s}- f_{i_t,j_t,k_t}||^2)$. 
If $\bc_i$ is the matching variable for the samples $\bx^s_i$ and $\bx^t_i$, 
$\bc_j$ for the samples $\bx^s_j$ and $\bx^t_j$ and $\bc_k$ for the samples $\bx^s_k$ and $\bx^t_k$, 
then $f_{i_s,j_s,k_s}$ is the feature consisting of the sine 
of the angles of the triangle formed by the data points $\bx^s_i,\bx^s_j$ 
and $\bx^s_k$ and $f_{i_t,j_t,k_t}$ consisting of the sine of the angles 
of the triangle formed by the data points $\bx^t_i,\bx^t_j$ and $\bx^t_k$. $\gamma$ is calculated from the mean pairwise squared distance between the features.

In addition to the graph-matching terms, 
we add a class-based regularization.
The group-lasso regularizer~\cite{friedman2010note} $\ell_2 - \ell_1$ 
norm term is equal to 
$f_g(\bC)=\sum_j\sum_{c}||[\bC]_{\mathcal{I}_{c}j}||_2$, 
where $||\cdot||_2$ is the $\ell_2$ norm and $\mathcal{I}_{c}$ 
contains the indices of rows of $\bC$ 
corresponding to the source-domain samples of class $c$. 
In other words, $[\bC]_{\mathcal{I}_{c}j}$ is a vector consisting of elements $[\bC]_{ij}$, 
where $i^{th}$ source sample belongs to class $c$ and the $j^{th}$ sample is in the target domain.
Minimizing this group-lasso term ensures that a target-domain sample only 
corresponds to the source-domain samples that have the same label. 

In the case of $n_t=n_s$, 
we have one-to-one matching between each source sample and each target sample. 
However, for the case $n_t \neq n_s$, we must allow multiple correspondences. 
Accordingly, if the constraint $\bC\mathbf{1}_{n_t}=\mathbf{1}_{n_s}$ ($\mathbf{1}_{n}$ is a $n \times 1$ vector of ones) implies that 
the sum of the correspondences of all the target samples to each source sample is one, 
then the second equality constraint $\bC^{T}\mathbf{1}_{n_s}=\frac{n_s}{n_t}\mathbf{1}_{n_t}$ implies 
that the sum of correspondences of all the source samples 
to each target sample should increase proportionately 
by $\frac{n_s}{n_t}$ to allow for the multiple correspondences. 
Hence, the overall optimization problem becomes \\[-0.1in]

\noindent
\emph{Problem UDA} 
\begin{align}
\label{prob1}
\hspace*{-3.0in}
 \min \limits_{\bC} f (\bC) ~ & = ~ \frac{1}{(n_sd)} ||\bC\bX^t-\bX^s||^2_\mathcal{F} +
 \lambda_2||\bC\bD^t - r\bD^s\bC||^2_\mathcal{F} \nonumber \\ &- \lambda_3\bH \otimes_1 \bc \otimes_2 \bc \otimes_3 \bc  +  \lambda_g\sum_j\sum_{c}||[\bC]_{\mathcal{I}_{c}j}||_2  \\
{\rm such~that} & ~ \bC \geq \bzero,  ~ \bC\mathbf{1}_{n_t} = \mathbf{1}_{n_s}, ~ {\rm and} ~\bC^T\mathbf{1}_{n_s}=(\frac{n_s}{n_t})\mathbf{1}_{n_t }.
\nonumber
\end{align}

\vspace*{-0.2in}
\subsection{Problem Solution}
Problem UDA can be solved quickly using second-order methods 
but has memory and computational issues related to storing the Hessian ($O(n_s^2n_t^2)$). 
Hence, first-order methods, specially conditional gradient 
method (CG)~\cite{frank1956algorithm} 
can be used to solve the problem. 
The CG method maintains the desirable structure of the solution such as sparsity required of $\bC$  by solving the successive linear minimization sub-problems 
over the constraint set~\cite{jaggi2013revisiting}. 
The linear programming (\emph{LP}) subproblem required to be solved 
is minimizing $\mathbf{Tr}(\bG^T\bC)$ ,$\bC \geq \mathbf{0}, \bC\mathbf{1}_{n_t}=\mathbf{1}_{n_s},\bC^T\mathbf{1}_{n_s}=\frac{n_s}{n_t}\mathbf{1}_{n_t}$, 
where $\bG$ is the gradient of the function $f$ in \emph{Problem UDA} and $\mathbf{Tr}(\cdot)$ is the trace operation. 
The gradient of each cost term can be derived as
\begin{align*}
\hspace*{-3.0in}
\nabla f_1(\bC) &= 2(\bC\bX^{t}-\bX^{s})\bX^{tT}/(n_sd),  \nonumber \\
\nabla f_2(\bC) &= ~2(\bC\bD^{t}\bD^{tT}-r\bD^{s}\bC\bD^{tT}-r\bD^{sT}\bC\bD^{t}+ r^{2}\bD^{sT}\bD^{s}\bC) ~,~\nonumber\\\nabla f_3(\bC)&=(\mathbf{[}\bH \otimes_1 \bc \otimes_2 \bc+\bH \otimes_1 \bc \otimes_3 \bc+\bH \otimes_2 \bc \otimes_3 \bc\mathbf{]}) 
~~\nonumber\\ \frac{\partial f_g}{\partial [\bC]_{ij}}&=\frac{[\bC]_{ij}}{||[\bC]_{\mathcal{I}_{c}(i)j}||_2}\cdot\delta(||[\bC]_{\mathcal{I}_{c}(i)j}||_2 \neq 0 ) \nonumber
\end{align*}
Here $[\cdot]$ operator on tensor term reshapes a vector 
into matrix of size $n_s \times n_t$. $c(i)$ is the class of the $i^{th}$ source sample. $\delta(\cdot)$ 
is an indicator function which is 1 if the argument is true and 0 otherwise. 
Thus, after obtaining the gradient 
$\bG=\nabla f(C)=\nabla f_1(C)+\lambda_2\nabla f_2(C)-
\lambda_3\nabla f_3(C)+\lambda_g\nabla f_g(C)$, we solve the linear minimization problem \emph{LP} mentioned  
before using the consensus form of ADMM~\cite{boyd2011admm}. 
We let $\ba=\mathbf{1}_{n_s}$ and $\bb=\frac{n_s}{n_t}\mathbf{1}_{n_t}$. 
Using the consensus ADMM form, we reformulate \emph{LP} as
\begin{align*}
\min & \left\{ g_1(\bC_1) +g_2(\bC_2)+g_3(\bC_3) \right\} \\
\text{such that } & \quad  \bZ=\bC_1=\bC_2=\bC_3
\end{align*}
where $g_1(\bC_1) =0.5\mathbf{Tr}(\bG^T\bC_1)+{I}(\bC_1\mathbf{1}_{n_t}=\ba)$,
$ g_2(\bC_2) =0.5\mathbf{Tr}(\bG^T\bC_2)+{I}(\bC_2^T\mathbf{1}_{n_s}=\bb)$ and $g_3(\bC_3) ={I}(\bC_3 \geq \mathbf{0})$. Here, $I(\cdot)$ is an indicator function which is 0 if argument is true and $\infty$ otherwise. $\mathbf{Z}$ is an intermediate variable to facilitate consensus ADMM.
Accordingly, the ADMM updates will be as follows
\begin{align*}\label{eq:admm1}
\bC_1^{k+1}=\underset{\bC_1\mathbf{1}_{n_t}=\ba}{\text{argmin}}(\mathbf{Tr}((0.5\bG+\bY_1^k)^T\bC_1)+\frac{\rho}{2}||\bC_1-\bZ^k||_F^2)\\
\bC_2^{k+1}=\underset{\bC_2^T\mathbf{1}_{n_s}=\bb}{\text{argmin}}(\mathbf{Tr}((0.5\bG+\bY_2^k)^T\bC_2)+\frac{\rho}{2}||\bC_2-\bZ^k||_F^2)\\
\bC_3^{k+1}=\Pi_{\{\bC|\bC\geq\mathbf{0}\}}(\bZ^k-\bY_3^k),\quad
\bZ^{k+1}=\frac{1}{3}(\sum_i^{}\bC_i^{k+1})\\
\bY_i^{k+1}=\bY_i^{k}+\bC_i^{k+1}-\bZ^{k+1} \quad \forall i=1,2,3
\end{align*}
Here $\bY_i$'s are dual variables. $\Pi$ is the projection operator. The penalty parameter is set $\rho=1$ without loss of generality 
since scaling $\rho$ is equivalent to scaling $\bG$. 
Updates for $\bC_1^{k+1}$ and $\bC_2^{k+1}$ are solved using Lagrange multiplier to obtain
\begin{align*}
\bC_1^{k+1}&=\bZ^k-\frac{\bG}{2}-\bY_1^{k}-\frac{1}{n_t}((\bZ^k-\frac{\bG}{2}-\bY_1^{k})\mathbf{1}_{n_t} -\ba)\mathbf{1}_{n_t}^T, \\
\bC_2^{k+1}&=\bZ^k-\frac{\bG}{2}-\bY_2^{k}-\frac{1}{n_s}\mathbf{1}_{n_s}(\mathbf{1}_{n_s}^T(\bZ^k
-\frac{\bG}{2}-\bY_2^{k}) -\bb^{T}), \\
 \bC_3^{k+1}&=\text{max}(\bZ^k-\bY_3^{k},\textbf{0})
\end{align*}
The ADMM updates are repeated for a fixed few-hundred iterations 
and the optimum value of \emph{LP} is set to final value of $\bZ$. 
This would complete one iteration of CG. 
After completing several such iterations of CG, 
we can obtain the optimal value of $\bC^{*}$. 
Using that, we can map the source domain data close to the target domain data 
using regression with $\bX^{s}$ and $\bC^{*}\bX^{t}$ as the input 
and output data respectively. 
The whole procedure can be repeated for a number of times at the end of which the adapted source data is used to train a classifier to be tested on target domain data.
The overall algorithm is given below in Algorithm 1.
\begin{algorithm}[]
\SetAlgoLined
 \textbf{Given :} Source Labelled Data $\bX^s$ and $\by^s$, and Target Unlabelled Data $\bX^t$\\
 \textbf{Parameters :} $\eta, \lambda_2, \lambda_3, \lambda_g, N_T$\\
 \textbf{Initialize :} $t_o=0$\\
 \textbf{Repeat}\\
 $\bX^s,\bX^t \leftarrow \mathbf{AP}(\bX^s,\bX^t,\eta)$, (Affinity Propagation)\\
  \textbf{Initialize:} $\bC_0 \in \mathcal{D}$, $t_i=1$\\
 \textbf{Repeat} (Conditional Gradient)\\
 \quad $\bG = \nabla_C f(\bC_0) $, $\bC_d \leftarrow \text{ADMM}(\bG,\ba.\bb)$\\
 \quad $\bC_1 = \bC_0 + \alpha(\bC_d - \bC_0), \quad \text{for} \quad \alpha = \frac{2}{t_i+2}$ \\
 \quad $\bC_0=\bC_1$ \quad \text{and} \quad $t_i=t_i+1$ \\
 \textbf{Until} Fixed Number of Iterations \\
 $\bC^{*} = \bC_0 $ and
Regress $\bM(\cdot) \quad \text{s.t.} \quad \bX^s \xrightarrow[]{\textbf{M}}\bC^*\bX^t$ \\
Map $\bX^s \leftarrow\bM(\bX^s)$ \quad and \quad $t_o=t_o+1$\\
 \textbf{Until} $t_o= N_T$ \\
 \textbf{Output :} Adapted Source Data $\bX^s$
\caption{UDA with Hyper-graph matching}
\end{algorithm}
\vspace*{-0.2in}

\subsection{Time and Space Complexity}
The AP clustering algorithm has a time complexity of ($O(n_s^2+n_t^2)$).
The ADMM updates are linear in the number of variables ($O(\eta^2n_sn_t)$) 
and they run for a fixed number of iterations. 
They are faster than interior point methods, 
which would result in cubic time-complexity in the number of variables. 
The overall time complexity will be multiplied by $N_T$. 
For the space complexity, the AP algorithm requires storing 
source and target similarity matrices of $O(n_s^2+n_t^2)$. 
Graph adjacency matrices also require space complexity of $O(\eta^2n_s^2+\eta^2n_t^2)$. 
For tensor storage, we use the sparse strategy~\cite{duchenne2011tensor} 
to obtain $O(m)$ space complexity, 
where $m$ is the number of non-zero entries and $m \propto \max(\eta n_s,\eta n_t)$ 

\section{EXPERIMENTAL RESULTS}
\begin{table*}[]
\centering
\caption{Comparing different methods in terms of classification accuracy (\%) of target data. 
Each task consists of \emph{source} $\rightarrow$ \emph{target}, 
where \emph{source} and \emph{target} represent any of the four domains: 
C(Caltech-256), A(Amazon), W(Webcam) and D(DSLR).}
\resizebox{18cm}{1.8cm}{
\label{table-1}
\begin{tabular}{|c|cccccccccccc|}
\hline
Method             & C $\rightarrow$ A & C $\rightarrow$ W & C $\rightarrow$ D & A $\rightarrow$ C & A $\rightarrow$ W & A $\rightarrow$ D & W $\rightarrow$ C & W $\rightarrow$ A & W $\rightarrow$ D & D $\rightarrow$ C & D $\rightarrow$ A & D $\rightarrow$ W \\
\hline
NA                 & 21.86                & 20.97                 & 22.73                 & 23.85                 & 24.31                 & 21.95                 & 19.03                 & 23.22                & 50.91                 & 23.79                 & 26.11                 & 51.67                    \\
GFK                & 33.41                 & 33.06                 & 35.19                 & 32.50                 & 33.89                 & 31.45                 & 25.12                 & 31.17                 & 79.22                 & 30.04                & 32.15                 & 71.87                   \\
SA                 & 34.12                 & 30.06                 & 33.11                 & 32.18                 & 32.56                 & 32.98                 & 30.15                 & 34.95                 & 68.31                 & 31.57                 & 34.25                 & 73.40                    \\
CORAL              & 31.76                 & 25.14                 & 27.40                 & 30.23                 & 28.33                 & 30.65                 & 24.92                 & 29.00                 & 78.05                 & 27.53                 & 28.95                 & 74.44                    \\
JDA                 & 40.56                 & 34.03                 & 34.28                 & 34.90                 & 34.58                 & 31.82                 & 32.63                 & 34.62                 & \textbf{85.19}                 & 30.50                 & 30.42                 & 78.89                     \\
OT    & \textbf{44.48}                 & 36.02                 & 36.88                 & \textbf{36.11}                 & 37.12                 & 39.35                 & 33.44                 & 37.36                 & 84.02                 & 31.82                 & 36.25                 & 79.23                     \\
Ours ($\eta=1$) & 42.30                 & \textbf{38.41}                 & \textbf{37.06}                 & 33.10                & \textbf{42.01}                 & \textbf{43.35}                 & \textbf{35.14}                & \textbf{40.71}                 & 83.14                 & \textbf{32.23}                 & \textbf{38.91}                 & \textbf{82.22}                   \\
Ours ($\eta=0.75$)  & 37.45                 & 30.56                 & 36.66                 & 34.20                 & 41.67                & 42.86                 & 30.23                 & 33.05                 & 77.97                 & 27.15                 & 33.81                 & 75.87                  \\
Ours ($\eta=0.5$)              & 34.73                 & 34.81                 & 33.77                & 33.63                 & 35.42                 & 37.66                 & 29.74                 & 35.24                 & 70.13                 & 24.61                 & 31.42                 & 66.67                \\
\hline    
\end{tabular}
}
\end{table*}
Our proposed method is tested against a standard dataset, 
known as the \textbf{Office-Caltech} dataset,
for domain adaptation of the object recognition task. 
It consists of a subset of the \textbf{Office} dataset~\cite{saenko2010adapting}.
This contains three different domains (Amazon, DSLR, Webcam) of images. 
The Amazon images are from the Amazon site, 
the DSLR images are captured with a high-resolution DSLR camera and
the Webcam domain contains images captured with a low-resolution webcam. 
The fourth domain consists of a subset of the object recognition dataset
\textbf{Caltech-256}~\cite{griffin2007caltech}. 
These four domains have ten common classes 
(Bike, BackPack, Calculator, Headphone, Keyboard, Laptop, 
Monitor, Mouse, Mug, Projector). 
Accordingly, we can obtain 12 DA task pairs.

The image features are the normalized SURF~\cite{surf} 
obtained as a 800-bin histogram. 
The classifier used was a 1-Nearest neighbor, as it is a hyper-parameter free. 
For our experiments, we considered a random selection of 20 samples per class  
(with the exception of 8 samples per class for the DSLR domain) for the source domain. 
One half of the target-domain data is used for domain adaptation. 
The accuracy is reported over 10 trials of the experiment. 
For our experiments, we used $\lambda_g=0.01$. 
We reported the best results obtained 
over the range $\lambda_2,\lambda_3 \in \{10^{-3},10^{-2}...10^{0}\}$ 
with a maximum $N_T=5$.  
For the transformation, we used a linear mapping 
with a $\ell_2$ regularization coefficient of $10^{-3}$. 
\begin{figure}[]
\centering
\includegraphics[width=6cm]{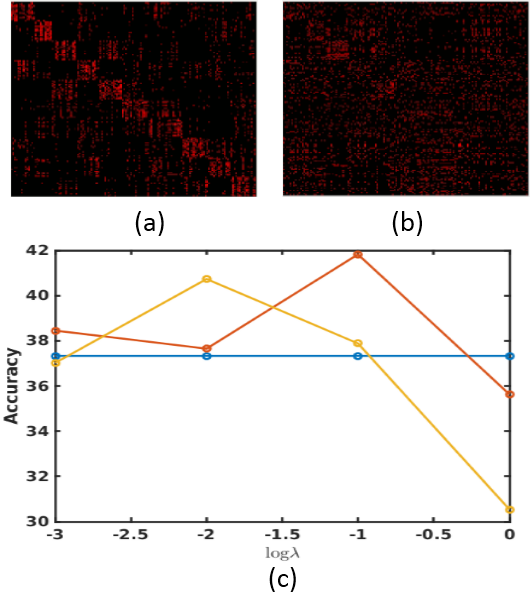}
\caption{Matching Matrix $\mathbf{C}$. (a) With $\lambda_g=0.01$. (b) With $\lambda_g=0$. (c) Accuracy values for different $\lambda_2, \lambda_3$. This is for $A \rightarrow W$ task.}
\label{sense}
\vspace*{-0.2in}
\end{figure}
We compared our approach against popular non-deep domain adaptation methods. Current methods involve deep architectures and would not compare fairly. So we include (a) The no adaptation baseline (NA), 
which consists of using the original classifier without adaptation; 
(b) Geodesic Flow Kernel (GFK)~\cite{gong2012geodesic}; 
(c) Subspace Alignment (SA)~\cite{fernando2013unsupervised};
(d) Joint Distribution Adaptation (JDA)~\cite{long2013transfer}, 
which jointly adapts both marginal and conditional distributions 
along with dimensionality reduction; 
(e) Correlation Alignment (CORAL)~\cite{sun2016return} and 
(f) Optimal Transport (OT)~\cite{courty2016optimal}.  

The comparison results are shown in Table~\ref{table-1}. 
We see that in almost all the cases, local DA methods like OT 
and our proposed method dominated over other global methods. 
Moreover, our method dominates over OT for most of the tasks. 
This is because our method exploits higher order 
structural similarity between the source and target data points. 
Also, in situations where OT is slightly better than our methods, 
the source and target data do not have enough structurally similar regions.

We also performed experiments to see how initial clustering 
affects recognition performance of our proposed method. 
We see the results in Table \ref{table-1} for $\eta = 0.75, 0.5$; 
that is, when the number of exemplars are around $75 \%, 50 \%$ 
of the total number of data-points, respectively. 
In general, the results showed decrease in performance 
with respect to $\eta=1$ but are still competitive 
with respect to some of the previous methods. 
The only exception is in the tasks $C \rightarrow W$, $A \rightarrow C$ and $W \rightarrow A$, 
where for decreasing $\eta$, the accuracy increases. 
This is because decreasing the number of exemplars 
decreases the possibility of including outliers as unwanted nodes 
in constructing the graph. 
Moreover, we analyzed the effect of regularization 
parameters $\lambda_2, \lambda_3, \lambda_g$ on the $A \rightarrow W$ task. 
Fig.~\ref{sense} (a) and (b) showed the matching 
matrix $\bC$ for $\lambda_g=0.01$ and $\lambda_g=0$, respectively. 
The rows and columns in the matrix $\bC$ represent the source and target, respectively. 
In Fig.~\ref{sense}(a), for $\lambda_g=0.01$, 
we see that 10 block-diagonal matrices representing 
all the 10 common categories appear as heatmap on the matrix $\bC$. 
It suggests that the group lasso regularizer 
allows discrimination of the classes more accurately compared to Fig.~\ref{sense}(b) 
that is with $\lambda_g=0$, where we have 
matching matrix $\bC$ with more uniform entries. 

In Fig.~\ref{sense} (c), 
we show the effect of varying $\lambda_2$ and $\lambda_3$. 
The blue straight line shows the accuracy result when $\lambda_2=\lambda_3=0$. 
The yellow line shows the accuracy result when $\lambda_2=0$ 
and $\log_{10}\lambda_3$ is varied. 
The red line shows the accuracy result when $\lambda_3=0$ 
and $\log_{10}\lambda_2$ is varied. 
The results suggest that including the higher order cost terms without weighing them heavily 
in the cost function improves performance over solely 
using the first-order matching. 

\section{CONCLUSION}
This paper proposed the use of hyper-graph matching 
between the source and target domains, 
which have previously not been used for unsupervised domain adaptation. 
We also proposed a computationally efficient optimization routine 
based on conditional gradient and ADMM. 
Results on object recognition dataset suggested our proposed UDA method 
to be competitive with respect to previous methods. 
In the future, we plan to extend our method to deep architectures 
that will learn the representations as well.

\bibliographystyle{IEEEbib}
\bibliography{TLMaster}

\end{document}